\documentclass[10pt,twocolumn,letterpaper]{article}

\usepackage{cvpr}
\usepackage{times}
\usepackage{epsfig}
\usepackage{graphicx}
\usepackage{amsmath}
\usepackage{amssymb}

\usepackage{graphicx}
\usepackage{booktabs}
\usepackage{subcaption}
\usepackage{amssymb}
\usepackage{multirow}
\usepackage[symbol]{footmisc}

\usepackage[margin=4pt,font=footnotesize,labelfont=bf,labelsep=endash,tableposition=top]{caption}

\usepackage[pagebackref=true,breaklinks=true,letterpaper=true,colorlinks,bookmarks=false]{hyperref}

\usepackage[breaklinks=true,bookmarks=false]{hyperref}

\cvprfinalcopy %

\def\cvprPaperID{****} %

\begin{document}

\title{CANet: Class-Agnostic Segmentation Networks with Iterative Refinement and Attentive Few-Shot Learning}
\author{Chi Zhang$^1$, ~~ Guosheng Lin$^1$\thanks{Guosheng Lin is the corresponding author.}, ~~ Fayao Liu$^2$, ~~ Rui Yao$^3$, ~~ Chunhua Shen$^2$\\
	$^{1}$Nanyang Technological University, Singapore\\
	$^{2}$The University of Adelaide, Australia\\
	$^{3}$China University of Mining and Technology, China\\
{ $\tt chi007@e.ntu.edu.sg$ }~~{ $\tt gslin@ntu.edu.sg$ }
}

\maketitle
\begin{abstract}
Recent progress in semantic segmentation is driven by deep Convolutional Neural Networks and large-scale labeled image datasets. However, 
data labeling for pixel-wise segmentation  is
tedious and costly. Moreover,  
a trained model can only make predictions within a set of pre-defined classes. In this paper, we present CANet, a class-agnostic segmentation network that performs few-shot segmentation on new classes with only a few annotated images available. Our network consists of a two-branch dense comparison module which performs multi-level feature comparison between the support image and the query image, and an iterative optimization module which iteratively refines the predicted results. Furthermore, we introduce an attention mechanism to effectively fuse information from multiple support examples under the setting of $k$-shot learning. Experiments on PASCAL VOC 2012 show that our method achieves  a mean Intersection-over-Union score of 55.4\% for 1-shot segmentation and 57.1\% for 5-shot segmentation,  outperforming   state-of-the-art methods by a large margin of 14.6\% and 13.2\%, respectively.
\end{abstract}
\section{Introduction}
Deep Convolutional Neural Networks have made significant breakthroughs in many visual understanding tasks including image classification~\cite{krizhevsky2012imagenet,resnet,simonyan2014very}, object detection~\cite{ren2015faster,he2017mask,redmon2016you}, and semantic segmentation~\cite{lin2017refinenet,chen2018deeplab,fcn}. One crucial reason is the availability of large-scale datasets such as ImageNet~\cite{imagenet} that enable the training of deep models. However, data labeling is expensive, particularly for dense prediction tasks, \eg, semantic segmentation and instance segmentation. In addition to that, after a model is trained, it is very difficult to apply the model to predict new classes. In contrast to machine learning algorithms, humans are able to segment a new concept from the image easily when only seeing a few examples. The gap between humans and machine learning algorithms motivates the study of few-shot learning that aims to learn a model which can be generalized well to new classes with scarce labeled training data.
\begin{figure}[]
\centering
\includegraphics[width=\linewidth]{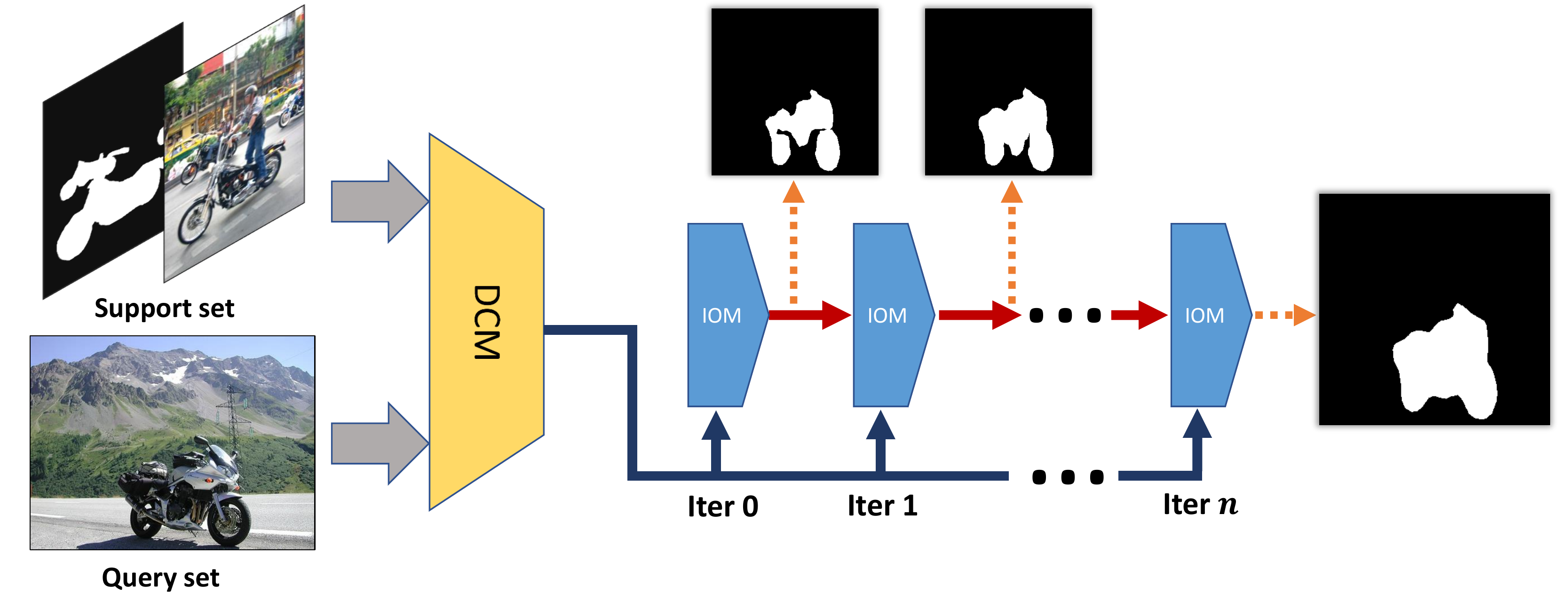}
\caption{Overview of our proposed network for 1-shot segmentation. Our framework consists of a dense comparison module (DCM) and an iterative optimization module (IOM). Given only one annotated training image, our network is able to segment  test images with new classes and iteratively optimize the results.}
\label{fig:pipline}
\vskip -1em
\end{figure}

In this paper, we undertake the task of few-shot semantic segmentation that only uses a few annotated training images to perform segmentation on new classes. Previous work~\cite{shaban2017one,rakelly2018conditional,Dong2018FewShotSS} on this task follows the design of two-branch structure which includes a support branch and a query branch. The support branch aims to extract information from the support set to guide segmentation in the query branch. We also adopt the two-branch design in our framework to solve the few-shot segmentation problem. 

Our network includes a two-branch dense comparison module, in which a shared feature extractor extracts representations from the query set and the support set for comparison. The design of the dense comparison module takes inspiration from metric learning~\cite{relation,snell2017prototypical} on image classification tasks where a distance function evaluates the similarity between images. However, different from image classification where each image has a label, image segmentation needs to make predictions on data with structured representation.
It is difficult to directly apply metric learning to dense prediction problems.  
To solve this, one straightforward approach is to make comparisons between all pairs of pixels. However, there are millions of pixels in an image and comparison of all pixel pairs takes enormous computational cost. Instead, we aim to acquire a global representation from the support image for comparison. Global image features prove  to be useful in segmentation tasks~\cite{liu2015parsenet,zhao2017pyramid,chen2017rethinking}, which can be easily achieved by global average pooling. Here, to only focus on the assigned category, we use global average pooling over the foreground area to filter out irrelevant information. Then the global feature is compared with each location in the query branch, which can be seen as a dense form of the metric learning approach.

Under the few-shot setting, the network should be able to handle new classes that are never seen during training. Thus we aim to mine transferable representations from CNNs for comparison. As is observed in feature visualization literature~\cite{zeiler2014visualizing,yosinski2015understanding}, features in lower layers relate to low-level cues, \eg, edges and colors while features in higher layers relate to object-level concepts such as categories. We focus on middle-level features that may constitute object parts shared by unseen classes. For example, if the CNN learns a feature that relates to \emph{wheel} when the model is trained on the class \emph{car}, such feature may also be useful for feature comparison on new vehicle classes, \eg, \emph{truck} and \emph{bus}. We extract multiple levels of representations in CNNs for dense comparison. 

As there exist variances in appearance within the same category, objects from the same class may only share a few similar features. Dense feature comparison is not enough to guide segmentation of the whole object area. Nevertheless, this gives an important clue of where the object is. In semi-automatic segmentation literature, weak annotations are given for class-agnostic segmentation, \eg, interactive segmentation with click or scribble annotations~\cite{interative,scribble} and instance segmentation with bounding box or extreme point priors~\cite{segevery,extreme}. Transferable knowledge to locate the object region is learned in the training process. Inspired by semi-automatic segmentation tasks, we hope to gradually differentiate the objects from the background given the dense comparison results as priors. We propose an iterative optimization module (IOM) that learns to iteratively refine the predicted results. The refinement is performed in a recurrent form that the dense comparison result and the predicted masks are sent to an IOM for optimization, and the output is sent to the next IOM recurrently. After a few iterations of refinement, our dense comparison module is able to generate fine-grained segmentation maps. Inside each IOM, we adopt residual connections to efficiently incorporate the predicted masks in the last iteration step.  Fig.~\ref{fig:pipline} shows an overview of our network for one-shot segmentation.

Previous methods for $k$-shot segmentation is based on the 1-shot model. They use non-learnable fusion methods to fuse individual 1-shot results, \eg, averaging 1-shot predictions or intermediate features. Instead, we adopt an attention mechanism to effectively fuse information from multiple support examples.

To further reduce the labeling efforts for few-shot segmentation, we explore a new test setting: our model uses the bounding box annotated support set to perform segmentation in the query image. We conduct comprehensive experiments on the PASCAL VOC 2012 dataset and COCO dataset to validate the effectiveness of our network. 
Main contributions of this paper are summarized as follows.
\begin{itemize}
\itemsep -0.1cm 
	\item We develop a novel two-branch dense comparison module which effectively exploits multiple levels of feature representations from CNNs to make dense feature comparison. 
	\item We propose an iterative optimization module to refine predicted results in an iterative manner. The ability of iterative refinement can be generalized to unseen classes with few-shot learning for generating fine-grained maps.
	\item We adopt an attention mechanism to effectively fuse information from multiple support examples in the $k$-shot setting, which outperforms non-learnable fusion methods of 1-shot results. 
	\item We demonstrate that given support set with weak annotations, \ie, bounding boxes, our model can still achieve comparable performance to the result with expensive pixel-level annotated support set, which further reduces the labeling efforts of new classes for few-shot segmentation significantly.
	\item Experiments on the PASCAL VOC 2012 dataset show that our method achieves a mean Intersection-over-Union score of 55.4\% for 1-shot segmentation and 57.1\% for 5-shot segmentation, which significantly outperform state-of-the-art results by 14.6\% and 13.2\%, respectively.
\end{itemize}

\section{Related Work}

\textbf{Semantic Segmentation.} Semantic segmentation is the task of classifying each pixel in an image to a set of pre-defined categories~\cite{lin2017refinenet,chen2018deeplab,fcn,lin2019refinenet,lin2016efficient}. State-of-the-art methods are based on Fully Convolutional Networks (FCNs), which often employ a convolutional neural network (CNN) pre-trained for classification  as the  backbone architecture. To fit the task of dense prediction, fully connected layers are replaced by a convolutional layer that predicts the label of each pixel. In order to capture abstract feature representations, CNNs adopt consecutive pooling operations or convolution striding to decrease the spatial resolution of feature maps. However, this conflicts with dense prediction tasks where the output should 
be of 
high resolution. In order to balance the output resolution and receptive field of %
the network,
dilated convolutions~\cite{chen2018deeplab} are often used in dense prediction tasks. Dilation removes downsampling operations in the last few layers and inserts holes to convolutional filters to enlarge the receptive field. In our model, we also adopt dilated convolutions to maintain spatial resolution. In fully supervised segmentation, training an FCN model requires a large number of expensive pixel-level annotated images, and once a model is trained, it can not perform segmentation on new categories. {\em Our model, on the other hand, can be generalized to any new categories with only a few annotated examples.
}

\textbf{Few-shot Learning.} Few-shot learning aims to learn transferable knowledge that can be generalized to new classes with scarce labeled training data. There exist many formulations on few-shot classification, including recurrent neural network with memories~\cite{santoro2016meta,munkhdalai2017meta}, learning to fine-tune models~\cite{finn2017model,ravi2016optimization}, network parameter prediction~\cite{bertinetto2016learning,wang2016learning}, and metric learning~\cite{snell2017prototypical,relation,koch2015siamese}. Metric learning based methods achieve state-of-the-art performance in the few-shot classification tasks and they have the trait of being fast and predicting in a feed-forward manner. Our work is most related to Relation Network~\cite{relation}. Relation Network meta-learns a deep distance metric to compare images and compute the similarity score for classification. The network consists of an embedding module which generates the representations of the images and a relation module that compares the embeddings and outputs a similarity score. Both modules are in the form of convolutional operations. The dense comparison module in our network can be seen as an extension of Relation Network in a dense form to tackle the task of segmentation.

\textbf{Few-shot Semantic Segmentation.} Previous work on few-shot semantic segmentation employs two-branch structures. Shaban \etal~\cite{shaban2017one} first adopt few-shot learning on semantic segmentation. The support branch directly predicts the weights of the last layer in the query branch for segmentation. In~\cite{rakelly2018conditional}, the support branch generates an embedding which is fused to the query branch as additional features. Our network also follows the two-branch design. However, different from previous work where two branches have different structures, the two branches in our network share the same backbone network. The models in previous methods focus on the 1-shot setting, and when extending 1-shot to $k$-shot, they apply 1-shot method independently to each support example and use non-learnable fusion methods to fuse individual predicted results at the image level or feature level. For example, Shaban \etal~\cite{shaban2017one} propose to use logic OR operation to fuse individual predicted masks and  Rakelly~\etal ~\cite{rakelly2018conditional} average the embedding in the support branch generated by different support examples. Instead, we adopt a learnable method through an attention mechanism to effectively fuse information from multiple support examples.

\section{Task Description}

Suppose that our model is trained on a dataset with the class set $\mathcal{C}_{train}$, our goal is to use the trained model to make the prediction on a different dataset with new classes $\mathcal{C}_{test}$ where only a few annotated examples are available. Intuitively, we train the model to have the ability that for a new class $c \not \in \mathcal{C}_{train}$, our model is able to segment the class from the images when only sees a few pictures of this class. Once the model is trained, the parameters are fixed and require no optimization when tested on a new dataset. 

We align training and testing with the episodic paradigm~\cite{vinyals2016matching} to handle the few-shot scenario. Specifically, given a $k$-shot learning task, each episode is constructed by sampling 1) a support (training) set $\mathcal{S} = \{(x_s^i,y_s^i(c))\}^{k}_{i=1}$, where $x_s^i \in \mathbb{R}^{H_i\times W_i\times 3}$ is an RGB image  and $y_s^i(c) \in \mathbb{R}^{H_i\times W_i}$ is a binary mask for class $c$ in the support image;  and 2) a query (test) set $\mathcal{Q}=\{x_q,y_q(c)\}$ where $x_q$ is the query image and $y_q(c)$ is the ground-truth mask for class $c$ in the query image. The input to the model is the support set $\mathcal{S}$ and the query image $x_q$, and the output is the predicted mask $\hat y_{q}(c)$ for class $c$ in the query image. As there may be multiple classes in one query image $x_q$, the ground truth query mask is different when a different label $c$ is assigned.  Fig.~\ref{fig:pipline} shows  an illustration of the task when $k = 1$.

\section{Method}
We propose a new framework that solves the few-shot semantic segmentation problem. We begin with the illustration of our model in the 1-shot setting first without loss of generality. Our network consists of two modules: the dense comparison module (DCM) and the iterative optimization module (IOM). The DCM performs dense feature comparison between  the support example and the query example, while IOM performs iterative refinement of predicted results. Fig.~\ref{fig:method} (a) shows an overview of our framework. To generalize our network from 1-shot learning to $k$-shot learning, we adopt an attention mechanism to fuse information from different support examples. Moreover, we propose a new test setting that uses support images with bounding box annotations for few-shot segmentation, which is described subsequently.

\subsection{Dense Comparison Module}
\begin{figure*}[]
	\centering
	\includegraphics[width=.8\linewidth]{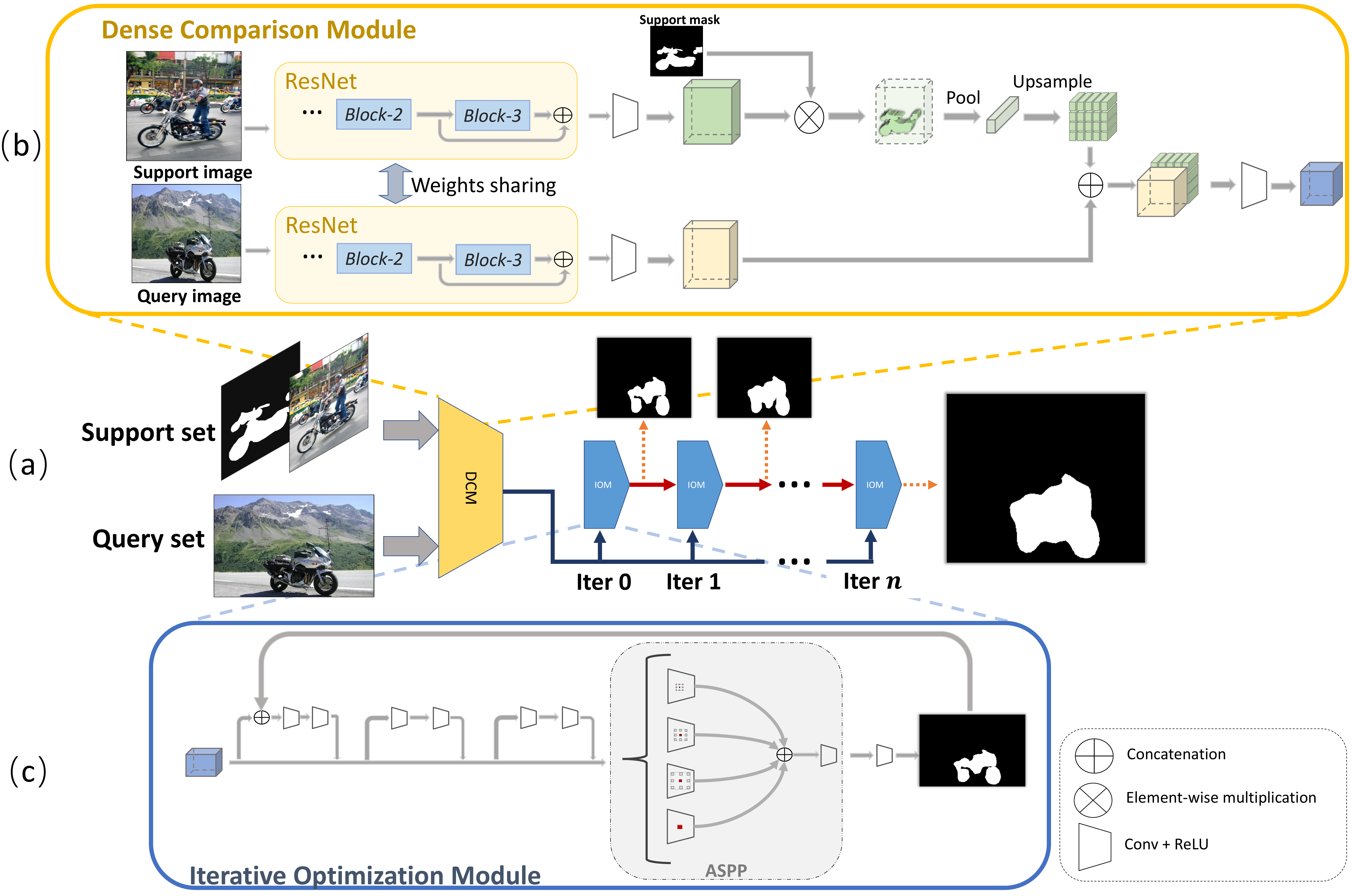}
	\caption{CANet for 1-shot semantic segmentation. (a) Overview of our network structure. (b) Dense Comparison Module. (c) Iterative Optimization Module. }
	\label{fig:method}
\vskip -1em
\end{figure*}

We develop a two-branch dense comparison module that densely compares each position in the query image with the support example, as shown in Fig.~\ref{fig:method} (b). The module consists of two sub-modules: a feature extractor that extracts representations and a comparison module that performs feature comparison.

\textbf{Feature Extractor.} The feature extractor aims to harvest different levels of representations from CNNs for feature matching. We use a ResNet-50~\cite{resnet} as the backbone of the feature extractor. As done in previous few-shot segmentation work, the backbone model is pre-trained on Imagenet~\cite{imagenet}. As is observed in CNN feature visualization literature~\cite{zeiler2014visualizing,yosinski2015understanding}, features in lower layers often relate to low-level cues, \eg, edges and colors while features in higher layers relate to object-level concepts such as object categories. In the few-shot scenario, our model should adapt to any unseen classes.  Thus we can not assume that a feature corresponding to an unseen category is learned during training. Instead, we focus on middle-level features that may constitute object parts shared by unseen classes. The layers in ResNet are divided into 4 blocks based on the spatial resolution which naturally correspond to 4 different levels of representation.  We choose features generated by $block2$ and $block3$ for feature comparison and abandon layers after $block3$. We use  dilated convolutions~\cite{chen2018deeplab} in layers after $block2$  to maintain the spatial resolution of feature maps. All feature maps after $block2$ have a fixed size of 1/8 of the input image.  Features after $block2$ and $block3$ are concatenated and encoded to 256 dimensions by $3 \times 3$ convolutions.  We investigate the choice of features for comparison in Section~\ref{section:ablation}. Both the support branch and the query branch use the same feature extractor. We keep the weights in ResNet fixed during training.

\textbf{Dense Comparison.} As there may be multiple object categories and cluttered backgrounds in the support image, we want to acquire an embedding that only corresponds to the target category for comparison. Here, we use global average pooling over the foreground area to squeeze the feature maps to a feature vector. Global image features turn out to be useful in segmentation tasks~\cite{liu2015parsenet,zhao2017pyramid,chen2017rethinking}, which can be easily achieved by global average pooling. In our network, we only average features over the foreground area to filter out irrelevant areas. After we obtain the global feature vector from the support set, we concatenate the vector with all spatial locations in the feature map generated by the query branch. This operation aims to compare all the spatial locations in the query branch to the global feature vector from the support branch. Then, the concatenated feature maps go through another convolutional block with 256 $3 \times 3 $ convolutional filters for comparison. 

For efficient implementation, we 
first bilinearly downsample the binary support mask to the same spatial size of the feature maps and then apply element-wise multiplication with the feature maps. As a result, features belonging to the background area become zero. Then we adopt global sum pooling and divide the resulting vector by the foreground area to obtain the average feature vector. We upsample the vector to the same spatial size of query features and concatenate them for dense comparison.

\begin{figure}[t]
	\centering
	\includegraphics[width=\linewidth]{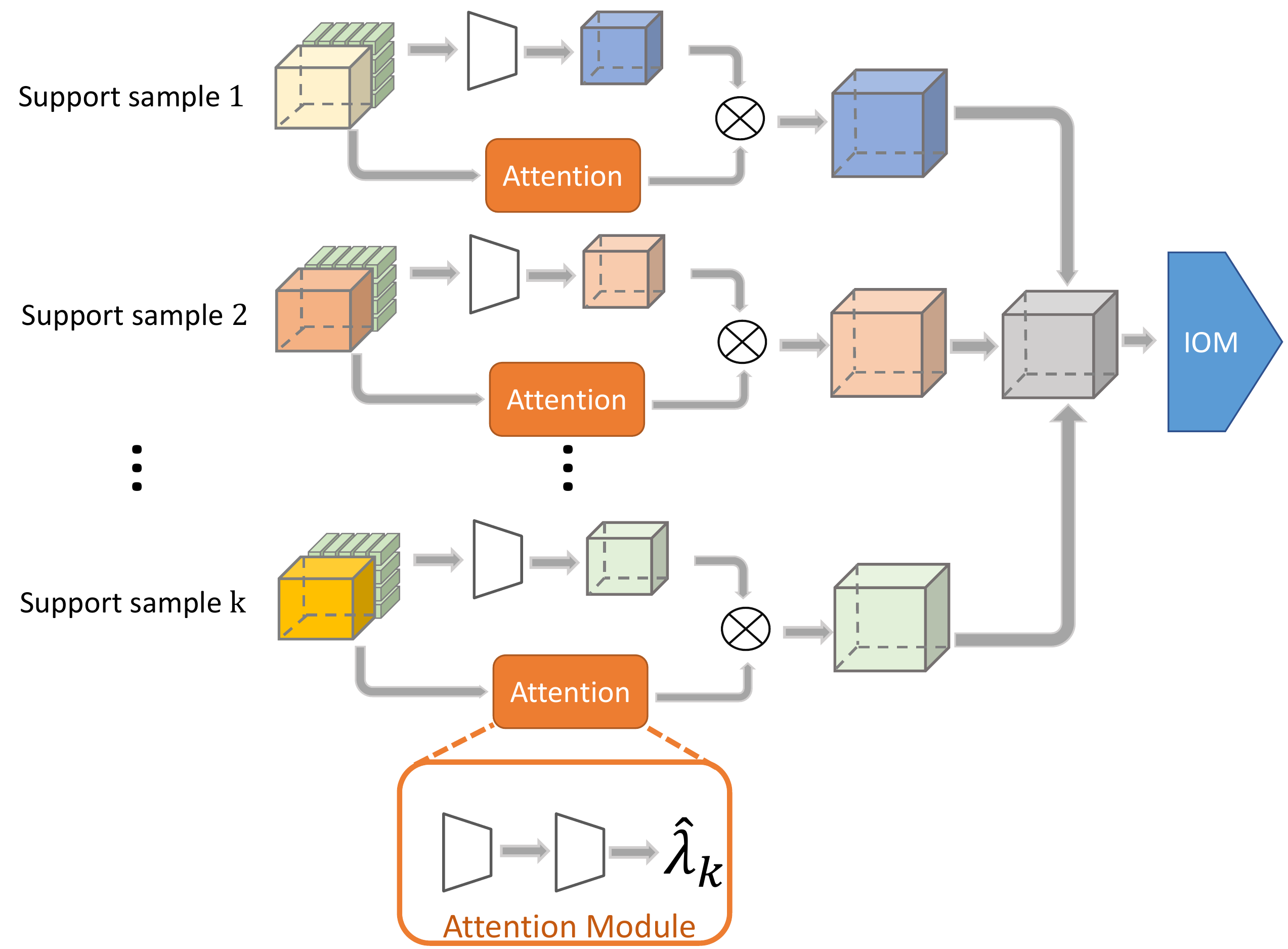}
	\caption{Attention mechanism for $k$-shot semantic segmentation. We use the softmax function to normalize the outputs of the attention module from different support examples.}
	\label{fig:attention}
\vskip -1em
\end{figure}

\subsection{Iterative Optimization Module}
As there exist variances in appearance within the same category, dense comparison can only match a part of the object, %
which may not be sufficiently powerful to accurately segment the whole object in the image.
We observe that the initial prediction is an important clue about the rough position of the objects. We propose an iterative optimization module to optimize the predicted results iteratively. The structure is shown in Fig.~\ref{fig:method} (c). The module's input is the feature maps generated by the dense comparison module and predicted masks from the last iteration. Directly concatenating feature maps with predicted masks as additional channels causes %
mismatch
to the feature distribution as there is no predicted mask for the first forward pass. Instead, we propose to incorporate the predicted masks in a residual form: 
\begin{equation}
M_t = x + F(x,y_{t-1}),
\end{equation}
where $ x $  is the output feature of the dense comparison module;  $y_{t-1}$ is the predicted masks from the last iteration step, and $M_t$ is the output of the residual block.  Function $F(\cdot )$ is the concatenation of feature $x$ and predicted masks $y_{t-1}$, followed by two $3 \times 3$ convolution blocks with 256 filters. Then we add two vanilla residual blocks with the same number of convolutional filters. On top of that, we use Atrous Spatial Pyramid Pooling module (ASPP) proposed in Deeplab V3~\cite{chen2017rethinking} to capture multi-scale information. The module consists of four parallel branches that include three $3 \times 3$ convolutions with atrous rates of 6, 12, and 18 respectively and a $1 \times 1$ convolution. The $1 \times 1$ convolution is operated on the image-level feature which is achieved by global average pooling. Then the resulting vector is bilinearly upsampled to the original spatial size. The output features from 4 branches are concatenated and fused by another $1 \times 1$ convolution with 256 filters. Finally, we use  $1\times 1$ convolution to generate the final masks which include a background mask and a foreground mask. We use a softmax function to normalize the score in each location, which outputs the confidence maps of the foreground and the background. The confidence maps are then fed to the next IOM for optimization. Our final result is achieved by bilinearly upsampling the confidence map to the same spatial size of the query image and classifying each location according to the confidence maps.  At the training time, to avoid the iterative optimization module over-fitting the predicted masks, we alternatively use predicted masks in the last epoch and empty masks as the input to IOM. The predicted masks $y_{t-1}$ is reset to empty masks with a probability of $p_r$.
This can be seen as dropout of the whole mask,
an extension of the standard dropout~\cite{srivastava2014dropout}.
In comparison to previous iterative refinement methods in segmentation literature~\cite{scribble,wang2018weakly,mcintosh2018recurrent}, our method integrates the refinement scheme into the model with residual connection so that the whole model could run in a feed-forward manner and is trained end-to-end.

\subsection{Attention Mechanism for k-shot Segmentation}
In order to efficiently merge information in the $k$-shot setting, we use an attention mechanism to fuse the comparison results generated by different support examples. Specifically, we add an attention module parallel to the dense comparison convolution in DCM (see Fig.~\ref{fig:attention}). The attention branch consists of two convolutional blocks. The first one has 256 $3 \times 3 $ filters, followed by $3 \times 3$ max pooling. The second one has one $3 \times 3$ convolution followed by a global average pooling. The result from the attention branch serves as the weight $\lambda$. Then, the weights from all support examples are normalized by a softmax function:
\begin{equation}
\hat{\lambda}_{i}=\frac{e^{\lambda_{i}}}{\sum_{j=1}^{k}e^{\lambda_{j}}}. 
\end{equation}
The final output is the weighted sum of features generated by different support samples.

\begin{figure}[t]
	\centering
	\includegraphics[width=0.8\linewidth]{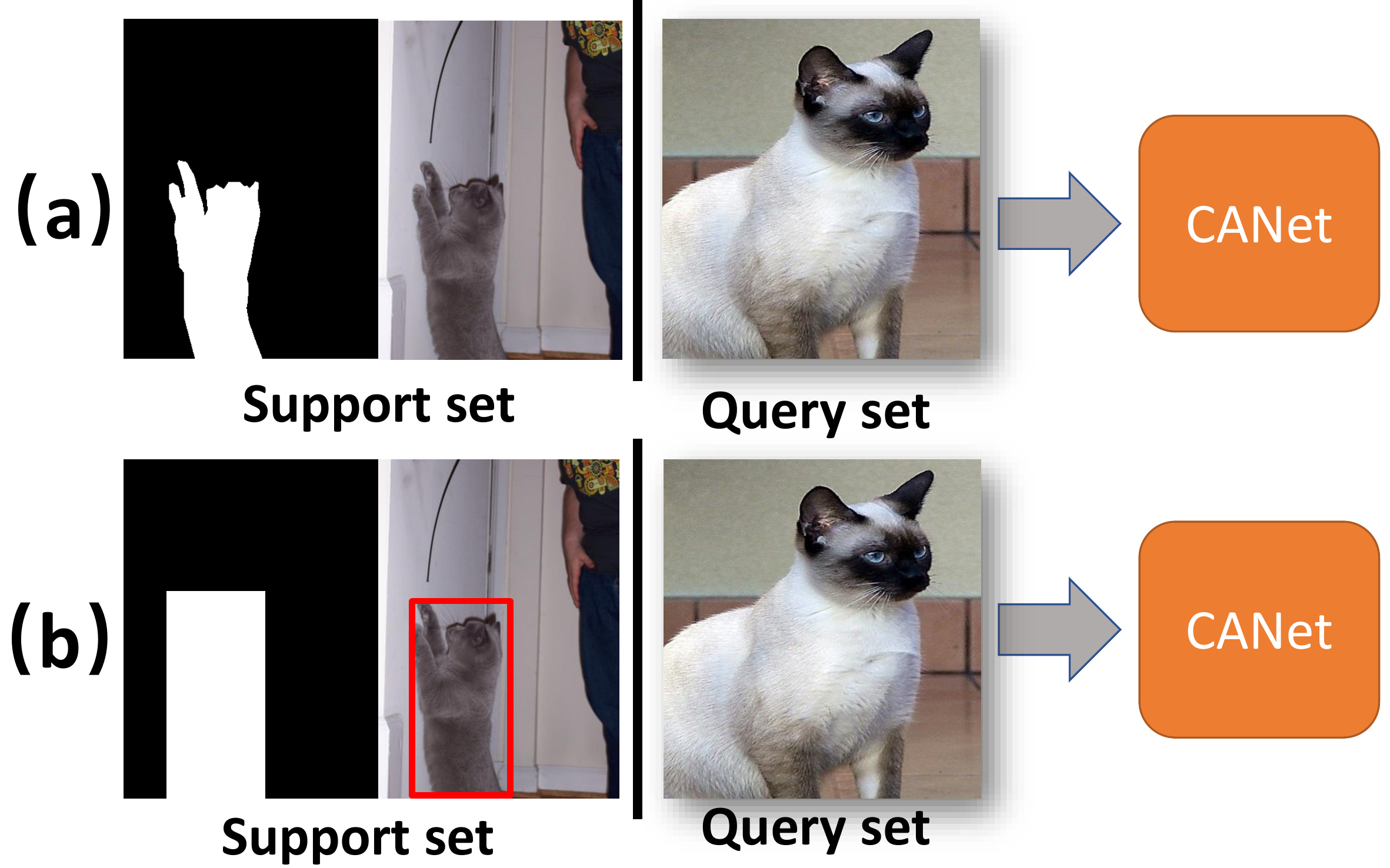}
	\caption{(a) CANet with pixel-wise annotated support set. (b) CANet with bounding box annotated support set. }
	\label{fig:boundingbox}
\vskip -1em
\end{figure}

\begin{table*}[t]
\small
\centering

\begin{subtable}{0.8\textwidth}
\small
\centering
\resizebox{0.81\textwidth}{!}{%
\begin{tabular}{lccccc|ccccc}
\hline
\multirow{2}{*}{Method} & \multicolumn{5}{c|}{1-shot} & \multicolumn{5}{c}{5-shot} \\ \cline{2-11} 
 & split-0 & split-1 & split-2 & split-3 & mean & split-0 & split-1 & split-2 & split-3 & mean \\ \hline\hline
OSLSM\cite{shaban2017one} & 33.6 & 55.3 & 40.9 & 33.5 & 40.8 & 35.9 & 58.1 & 42.7 & 39.1 & 43.9 \\\hline
CANet & \textbf{52.5} & \textbf{65.9} & \textbf{51.3} & \textbf{51.9} & \textbf{55.4} & \textbf{55.5} & \textbf{67.8} & \textbf{51.9} & \textbf{53.2} & \textbf{57.1} \\ \hline
\end{tabular}%
}
\caption{1-shot and 5-shot results under the meanIoU evaluation metric.}
\end{subtable}

\medskip%
\begin{subtable}{0.8\textwidth}
\small
\centering
\resizebox{0.81\textwidth}{!}{%
\begin{tabular}{lccccc|ccccc}
\hline
\multirow{2}{*}{Method} & \multicolumn{5}{c|}{1-shot} & \multicolumn{5}{c}{5-shot} \\ \cline{2-11} 
 & split-0 & split-1 & split-2 & split-3 & mean & split-0 & split-1 & split-2 & split-3 & mean \\ \hline\hline
OSLSM~\cite{shaban2017one} & - & - & - & - & 61.3 & - & - & - & - & 61.5 \\
co-FCN~\cite{rakelly2018conditional} & - & - & - & - & 60.1 & - & - & - & - & 60.2 \\
PL~\cite{Dong2018FewShotSS} & - & - & - & - & 61.2 & - & - & - & - & 62.3 \\\hline
CANet & \textbf{71.0} & \textbf{76.7} & \textbf{54.0} & \textbf{67.2} & \textbf{66.2} & \textbf{74.2} & \textbf{80.3} & \textbf{57.0} & \textbf{66.8} & \textbf{69.6} \\ \hline
\end{tabular}%
}
\caption{1-shot and 5-shot results under the FB-IoU evaluation metric.}
\end{subtable}

\caption{Results on the PASCAL-5$^i$ dataset. Our proposed method outperforms all previous methods under both evaluation metrics and sets a new state-of-the-art performance (bold).   }
\label{pascal}
\vskip -1em
\end{table*}

\subsection{Bounding Box Annotations}
As the essence of our dense comparison module is to densely compare each location in the query image to the global representation provided by the support example, we explore a new form of support set annotation that uses bounding boxes. Compared with pixel-wise annotations, the bounding box annotation uses a rectangular box to denote the object area, which is often used in object detection tasks. Labeling bounding box annotations is much cheaper than pixel-wise labeling. We relax the support set by treating the whole bounding box area as the foreground. We test our model under this setting to evaluate the capability of our framework. The comparison of the two test settings is shown in Fig.~\ref{fig:boundingbox}.

\section{Experiments}

To evaluate the performance of our proposed method, we conduct extensive experiments on the PASCAL VOC 2012 dataset and COCO dataset. Our network is trained end-to-end. The loss function is the mean of cross-entropy loss over all spatial locations in the output map. Our network is trained using SGD for 200 epochs with the PyTorch library on Nvidia Tesla P100 GPUs. We set the learning rate to 0.0025 and set probability $p_r$ to 0.7. We use a mini-batch of 4 episodes for training on  PASCAL-$5^i$ and 8 on COCO. At inference time, we iteratively optimize the predicted results for 4 times after the initial prediction. 

\textbf{Evaluation Metric.} There is a minor difference of evaluation metrics in previous work. Shaban \etal~\cite{shaban2017one} measure the per-class foreground Intersection-over-Union (IoU) and use the average IoU over all classes (meanIoU) to report the results. While in~\cite{rakelly2018conditional,Dong2018FewShotSS}, they ignore the image categories and calculate the mean of foreground IoU and background IoU over all test images (FB-IoU). We choose the meanIoU evaluation metric for our analysis experiments due to the following reasons: 1) The numbers of test samples in different classes are not balanced (\eg, 49 of class \emph{sheep} \emph{vs.} 378 of class \emph{person}). Ignoring the image categories may lead to a biased result towards the class with more images. Also, we can observe the effectiveness of our model in different classes with the meanIoU evaluation metric. %
2) As most objects are small relative to the whole image, even though the model fails to segment any objects, the background IoU can still be very high, thus failing to reflect the capability of the model. 
3) Foreground IoU is more often used in binary segmentation literature (\eg, video segmentation and interactive segmentation). Nevertheless, we still compare our results with previous work under both evaluation metrics.

\subsection{PASCAL-5$^i$}
PASCAL-5$^i$ is a dataset for few-shot semantic segmentation proposed in~\cite{shaban2017one}. It is built on images from PASCAL VOC 2012 and extra annotations from SDS~\cite{hariharan2014simultaneous}. 20 object categories from PASCAL VOC are evenly divided into 4 splits with three splits for training and one split for testing. At test time, 1000 support-query pairs are randomly sampled in the test split. More details of PASCAL-5$^i$ can be found in~\cite{shaban2017one}.

\subsubsection{Comparison with the State-of-the-art Methods}
We compare our model with the state-of-the-art methods in Table~\ref{pascal}. Table~\ref{pascal} (a) shows the results evaluated under the meanIoU evaluation metric and Table~\ref{pascal} (b) shows the results under the FB-IoU metric. For the performance of~\cite{shaban2017one} under the FB-IoU metric, we quote the result reproduced in~\cite{rakelly2018conditional}. {\em Our model significantly outperforms the state-of-the-art methods under both evaluation metrics}. Particularly, our meanIoU score outperforms the state-of-the-art results by 14.6\% for the 1-shot task and 13.2\% for the 5-shot task.

\begin{figure*}[t]
\begin{center}
\includegraphics[width=\linewidth]{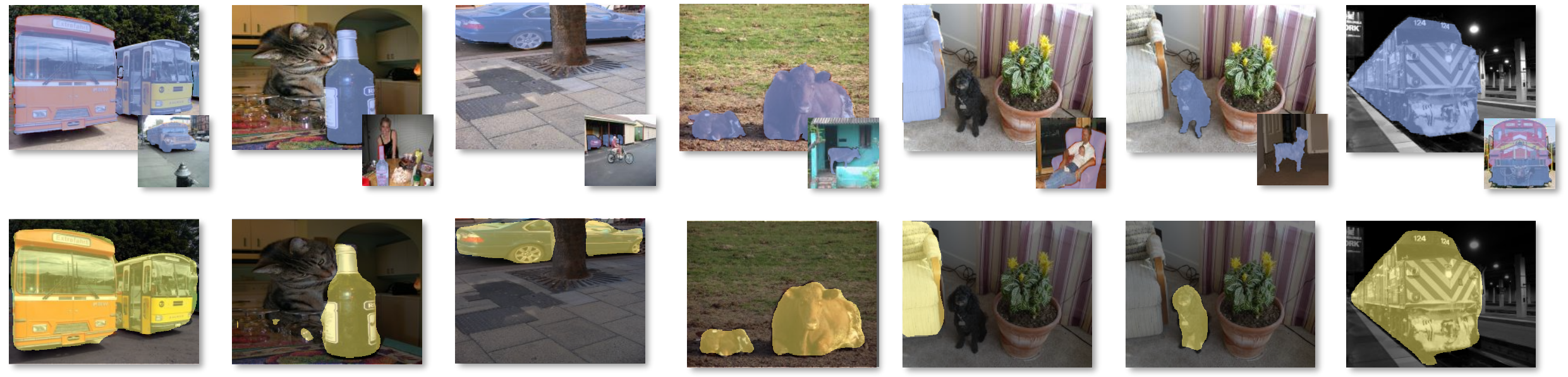}
	\caption{Qualitative examples of 1-shot segmentation on the PASCAL-5$^i$ dataset. The first row is query images and support images (right bottom) with ground-truth annotations. The second row is our predicted results. Note that the 5th and the 6th examples have the same query images and our model is able to segment different classes when different support examples are presented. }
	\label{pascalimg}
\end{center}
\vskip -1em
\end{figure*}

\textbf{Qualitative Results.} Fig.~\ref{pascalimg} shows some qualitative examples of our segmentation results. Note that given the same query image, our model is able to segment different classes when different support examples are presented (See the 5th and the 6th examples in Fig.~\ref{pascalimg}).

\subsubsection{Experiments on Bounding Box Annotations}
We evaluate CANet with the bounding box annotated support set at test time. We acquire bounding box annotations from the PASCAL VOC 2012 dataset and SDS~\cite{hariharan2014simultaneous}. The support mask is the region inside the bounding box of one instance instead of all instances in a support image. The instance is chosen randomly. As is shown in Table~\ref{table:box}, {\em the performance with bounding box annotated support set  is comparable to the result with expensive pixel-level annotated support set}, which means our dense comparison module is able to withstand noise introduced by the background area within the bounding box.

\begin{table}[t]
\small
\centering
\begin{tabular}{lc}
\hline
\multicolumn{1}{c}{Annotation}   & Result (meanIoU \%) \\ \hline\hline
Pixel-wise labels& 54.0                  \\
Bounding box  & 52.0                \\ \hline
\end{tabular}%
\caption{Evaluation with different support set annotations. Our model with bounding box annotated support set can achieve comparable performance to the result with pixel-wise annotations}
\label{table:box}
\vskip -1em
\end{table}

\subsubsection{Ablation Study}
\label{section:ablation}
We implement extensive ablation experiments on the PASCAL-5$^i$ dataset to inspect the effectiveness of different components in our network. All results are average meanIoU over 4 splits on the PASCAL-5$^i$ dataset.

\textbf{Features for Comparison.}  In Table~\ref{ablation:features}, we compare our model variants that use different levels of feature in ResNet-50 for feature comparison. In all cases, we encode the features to 256 dimensions before comparison and we do not adopt iterative optimization. We experiment feature comparison with single block and multiple blocks. When single block is used for comparison, \emph{block3} performs the best. When multiple blocks are used for comparison, the combination of \emph{block2} and \emph{block3} achieves the best result.  The reason is that \emph{block2} corresponds to relatively low-level cues, which alone is not enough to match object parts. While \emph{block4} corresponds to high-level features, \eg, categories, and incorporates a great number of parameters (2048 channels), which makes it hard to optimize under the few-shot setting. The combination of \emph{block2} and {block3} is the best for matching class-agnostic object parts. 

We also implement experiments with VGG16 as the feature extractor. We choose features of stage 2, 3, and 4 (out of 5). The final multi-scale test result  with VGG as the backbone is 54.3\%. Compared with the ResNet50 version (55.4\%), the performance only drops by 1.1\% and still significantly outperform the state-of-the-art results.

\begin{table}[t]
\centering
\resizebox{0.30\textwidth}{!}{%
\begin{tabular}{cccc}
\hline
$block2$ & $block3$ & $block4$ & meanIoU \\ \hline\hline
\checkmark     &        &        &    46.6     \\
       & \checkmark      &        &     50.8    \\
       &        & \checkmark      &    48.4     \\
\checkmark      & \checkmark      &        &    \textbf{51.2}     \\
       & \checkmark      & \checkmark      &    49.2     \\
\checkmark      &        & \checkmark      &    49.6     \\
\checkmark      & \checkmark      & \checkmark      &    49.5     \\ \hline
\end{tabular}%
}
\caption{Ablation experiments on the choice of features in ResNet for comparison. The combination of features after \emph{block2} and \emph{block3} achieves the best result. }
\label{ablation:features}
\vskip -1em
\end{table}

\begin{figure}
\begin{center}
\includegraphics[width=1\linewidth]{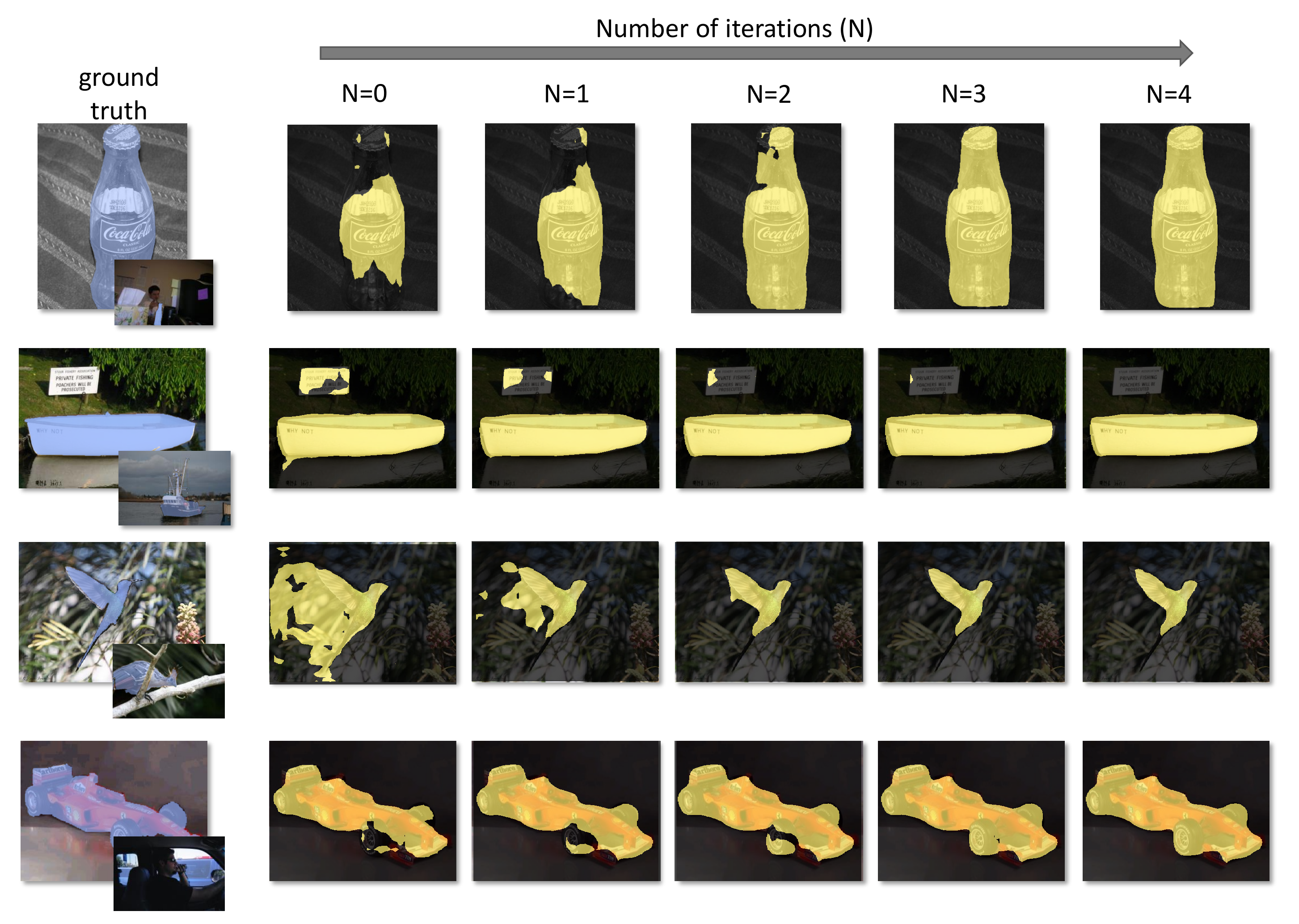}
	\caption{Visualization of the iterative optimization process. The first column shows the query and support images with ground-truth masks annotated. The rest columns show our iterative optimization results.    }
	\label{fig:iter}
\end{center}
\vskip -1em
\end{figure}

\textbf{Iterative Optimization Module.} To validate the effectiveness of our proposed iterative optimization module, we compare our network with a baseline model that does not employ additional IOM for optimization, \ie, the initial prediction from CANet(CANet-Init). We also compare our iterative optimization scheme with DenseCRF \cite{crf}, which is a post-processing method widely used in segmentation literature to refine segmentation maps. Table~\ref{table:iter} shows the results of different model variants. As is shown, the iterative optimization yields 2.8\% improvement over the initial prediction. DenseCRF does not significantly improve the few-shot segmentation prediction. We visualize the results and find that for the predicted masks which successfully locate most of the object region, DenseCRF can effectively improve segmentation results, particularly in the region of object boundaries. However, for failure masks, \eg, false localization of objects, DenseCRF expands false positive regions, which deteriorates the IoU score. Our IOM, on the other hand, can effectively fill the object region and remove irrelevant areas in a learnable way. We visualize the intermediate results of our iterative optimization process in Fig.~\ref{fig:iter}.

\begin{table}[t]
\small 
\centering
\begin{tabular}{lc}
\hline
\multicolumn{1}{c}{Method} & Result (meanIoU \%) \\ \hline\hline
CANet-Init  & 51.2                  \\ 
CANet-Init + DenseCRF & 51.9                 \\ \hline
CANet & \textbf{54.0}                  \\ 
\hline
\end{tabular}%
\caption{Ablation experiments on the iterative optimization module. CANet-Init denotes the initial prediction from CANet without additional optimization. Our iterative optimization scheme outperforms the baseline models by 2.8\% and is more effective in refining the segmentation maps than DenseCRF.}
\label{table:iter}
\vskip -1em
\end{table}

\textbf{Attention \emph{vs.} Feature Fusion \emph{vs.} Mask Fusion. } In the $k$-shot setting, we compare our attention mechanism to several solutions in previous work: 1) Feature-level average fusion. We experiment the method in~\cite{rakelly2018conditional}, which is to average the features generated by different support examples. 2) Logic OR fusion for masks. Shaban \etal~\cite{shaban2017one} use 1-shot model to make predictions with each support example and use logic OR operation to fuse individual predicted masks. Logic OR operation means that a position is predicted as foreground if any support example predicts it as foreground. 3) Average fusion for masks. Moreover, we also experiment with average operation to fuse individual 1-shot predicted confidence maps. We report the results of CANet with different fusion solutions in  Table~\ref{table:attention}. Our attention mechanism performs the best and brings the most increment over 1-shot baseline. This indicates that a learned attention module can be more effective in fusing information from different support examples than non-learnable fusion methods in feature level or image level. Using logic OR operation to fuse predicted masks does not show improvement over the 1-shot result.

\begin{table}[t]
\small
\centering
\begin{tabular}{lcc}
\hline
\multicolumn{1}{c}{Method}      & Result (meanIoU \%) & Increment \\ \hline\hline
1-shot baseline &54.0 & 0 \\ \hline
Feature-Avg & 55.0               & 1.0         \\
Mask-Avg    & 54.5               & 0.5         \\
Mask-OR     & 53.4               & -0.6         \\\hline
Attention   & \textbf{55.8}               & \textbf{1.8}         \\ \hline
\end{tabular}%
\caption{Comparison of different 5-shot solutions. Our attention method performs the best and brings the most increment in the meanIoU score over the 1-shot baseline. }
\label{table:attention}
\vskip -1em
\end{table}

\textbf{Multi-scale Evaluation.}
We also experiment multi-scale evaluation as is commonly done in segmentation literature. Specifically, we re-scale the query image by [0.7, 1, 1.3 ] and average their predicted results. Multi-scale evaluation brings 1.4\% and 1.3\% meanIoU improvement in 1-shot and 5-shot settings, respectively.

\subsection{COCO}
COCO 2014~\cite{coco} is a challenging large-scale dataset, which contains 80 object categories. The original dataset contains 82,783 and 40,504 images for training and validation respectively. Directly experimenting on the original dataset is very demanding on time and computation. Instead, we choose a subset of the original dataset to evaluate our model and for further research on this topic. We choose 40 classes for training, 20 for validation and 20 for test, which contain 39,107 (train), 5,895 (validation) and 9,673 (test) samples, respectively.  Training images are chosen from the COCO training set, while validation and test images are chosen from the COCO validation set.

For the 1-shot task, we compare our network with the baseline model that does not employ additional iterative optimization (CANet-Init), and for the 5-shot task, we compare our attention mechanism with three non-learnable fusion methods described in Section~\ref{section:ablation}. The result is shown in Table~\ref{tab:coco}. In the 1-shot setting, our iterative optimization scheme brings  4.1\% meanIoU improvement. Multi-scale evaluation shows extra 3.3\% increase. In the 5-shot setting, our attention mechanism outperforms all non-learnable methods. Multi-scale evaluation obtains another 1.9\% gain.

\begin{table}[t]
\small
\centering

\begin{subtable}{0.45\textwidth}
\centering
\begin{tabular}{lcc}
\hline
\multicolumn{1}{c}{Method} &MS& Result (meanIoU \%) \\ \hline\hline
CANet-Init    & & 42.2                  \\
CANet &          &46.3                  \\ 
CANet &  \checkmark &\textbf{49.9}                  \\ \hline
\end{tabular}%
\caption{1-shot results on COCO dataset.}
\end{subtable}

\medskip%
\begin{subtable}{0.45\textwidth}
\centering
\begin{tabular}{lcc}
\hline
\multicolumn{1}{c}{Method} & MS & Result (meanIoU \%) \\ \hline\hline
Feature-Avg   &  & 48.9                  \\
Mask-Avg &       &   49.2                  \\ 
Mask-OR &      & 46.2               \\\hline
Attention &  &49.7              \\
Attention  & \checkmark  & \textbf{51.6}                  \\ \hline
\end{tabular}%
\caption{5-shot results on COCO dataset.}
\end{subtable}

\caption{MeanIoU results on COCO dataset. MS denotes multi-scale evaluation. }
\label{tab:coco}
\vskip -1em
\end{table}
\section{Conclusion}
We have presented CANet, a novel class-agnostic segmentation network with few-shot learning. The dense comparison module exploits multiple levels of feature in CNNs to perform dense feature comparison and the iterative optimization module learns to iteratively refines the predicted results. Our attention mechanism for solving the $k$-shot problem turns out to be more effective than non-learnable methods. Comprehensive experiments show the effectiveness of our framework, and the performance significantly outperforms all previous work. 

\section{Acknowledgements} G. Lin's participation was partly supported by the National Research Foundation Singapore under its AI Singapore Programme [AISG-RP-2018-003] and a MOE Tier-1 research grant [RG126/17 (S)]. R. Yao's participation was supported by the National Natural Scientific Foundation of China (NSFC) under Grant No. 61772530. We would like to thank NVIDIA for GPU donation. 

{\small
\bibliographystyle{ieee}
\bibliography{egpaper_final}
}

\end{document}